\documentclass{article}

\usepackage{microtype}
\usepackage{graphicx}
\usepackage{subcaption}
\usepackage{booktabs}
\usepackage{hyperref}

\usepackage[preprint]{icml2026}

\usepackage{amsmath}
\usepackage{amssymb}
\usepackage{mathtools}
\usepackage{amsthm}
\usepackage[capitalize,noabbrev]{cleveref}

\theoremstyle{definition}

\icmltitlerunning{Assessing LLMs' mathematical abilities}

\begin{document}

\twocolumn[
  \icmltitle{Assessing LLMs' mathematical abilities requires understanding the
    various mechanisms of mathematical creativity}

  \begin{icmlauthorlist}
    \icmlauthor{Silvère Gangloff}{irafm}
  \end{icmlauthorlist}
  \icmlaffiliation{irafm}{IRAFM, University of Ostrava, Ostrava, Czech Republic}
  \icmlcorrespondingauthor{Silvère Gangloff}{silvere.f.gangloff@gmail.com}

  \icmlkeywords{Machine Learning, mathematics, mathematical creativity, LLMs,
    philosophy of mathematics}

  \vskip 0.3in
]

\printAffiliationsAndNotice{}

\begin{abstract}
How should we assess whether large language models can perform mathematical invention? I
argue that this question is currently underspecified: mathematical creativity is not one
capacity but several mechanistically distinct modes of meaning-making - reflexive
introspection on mathematical practice, analogical import from the sciences, problem-driven
construction, and the bridging of
distant domains - together with a further, cross-cutting distinction between meaning
pursued because a pattern was observed and meaning pursued because it is strategically
wanted, a distinction I develop through the case of conjecture-formation. These
mechanisms are likely non-substitutable, so that competence in one does not transfer to
the others. Grounding each in a historical case study and in an architecture-level
account of current transformer-based systems, I suggest that today's models concentrate
their competence in modes shaped by recombination and search over existing building
blocks; if that description holds, the remaining modes are out of reach in principle, not
just slower - though whether it holds is itself the open, empirical part. Because proof
is getting cheaper as AI improves at generating it - a shift the field's own leading
voices are now diagnosing - mathematical value is migrating toward the modes current
systems cannot yet perform, and evaluations of AI mathematical ability should be
organized around this taxonomy rather than around aggregate benchmarks that conflate it.
\end{abstract}

\section{Introduction}
\label{sec:intro}

G.~H.~Hardy opens \emph{A Mathematician's Apology} by declaring that a mathematician,
like a painter or a poet, is \textit{``a maker of patterns,''} and that it is the beauty of an
idea, not the rigor of its proof, that is the true measure of the
work~\citep{hardy1940apology}. Yet the working life of the same Hardy tells a different
story: for years, Hardy and Littlewood devoted themselves to supplying rigorous proof for
claims that Ramanujan had reached by an intuition neither of them could fully reconstruct,
precisely because intuition alone, however striking, was not treated by the field as
sufficient grounds for trusting a claim. Hardy is therefore a witness for this paper's
position even while representing the era that made proof mathematics' primary currency:
proof's dominance was plausibly never because it \emph{is} the value of mathematics, but
the only scalable technology for checking an intuition. When verification becomes
available by other means, the true scarce resource - whose intuition to trust, what is
worth the labor of
formalizing - is exposed again.

That shift is no longer speculative. In his 2026 ICM plenary, \textit{``Mathematics in the Age of
AI,''} Terence Tao argues that the field is moving from a shortage of proofs to an
overwhelming abundance of them, and warns explicitly that \textit{``the inherently ungrounded
nature of generative AI, combined with the financial incentives of AI companies, makes
the use of these tools particularly vulnerable to''} Goodhart's law - the tendency of a
measure, once made a target, to stop being a good measure~\citep{tao2026mathematics}. Tao
decomposes mathematical work into five stages beyond initial problem-solving: proof
generation, verification, exposition, publication, and canonicalization for future
generations, observing that AI is already strong at generation and improving quickly at
verification, while the remaining stages still require sustained human involvement.

Let's be more precise about what this five-stage account does and does not claim, since
it is easy to mistake for a competing version of the taxonomy developed here.
Tao's stages are a \emph{sequential, lifecycle} account of what happens to a piece of
mathematics once it already exists as a claim: generated as a proof, checked, published,
absorbed into the literature and curriculum. Every stage presupposes a claim is already in
hand. This paper asks a prior question: where does the content - for instance,
concepts, conjectures and axioms - come from in the first place? Tao folds
that question silently into \textit{``problem-solving,''} since his interest is in what happens to
content after it exists, not in its genesis. Turing's 1936 paper illustrates the
difference: its results pass through generation, verification, and canonicalization into
the computer science curriculum exactly as Tao's stages describe, but none of those
stages touch the prior, reflexive move that made the paper possible - noticing that a
human clerk mechanically following a rulebook could itself become a formal
object~\citep{turing1936computable}. That move precedes stage one entirely. The two
taxonomies are not competing slices of the same phenomenon: one classifies the lifecycle
of mathematical content, the other the generative mechanism that produces it.

This paper is a direct response to an open question left by the closest existing work in
this vein. \citet{zahavy2026jump} uses Peirce's classical division of inference into
deduction, induction, and abduction to argue that while large language models have
mastered the deductive verification of theorems and the inductive compression of data,
they lack the abductive \textit{``jump''} from sense experience to formal axioms that he takes
Einstein's discovery of general relativity to exemplify, and he proposes that physically
grounded, actionable world models are the missing piece. I take his argument seriously
and build directly on its Peircean apparatus; I do not think it fails so much as it stops
exactly where it says it does. Its conclusion explicitly, and I think correctly, declines
to force a physics-derived axis onto a domain he had not examined, noting only that in
mathematics \textit{``[sense experience] may be grounded in high-dimensional topology or have
other goals such as generality or minimality,''} without developing the point further.
This paper departs from that single, undeveloped sentence. His account treats the
missing capacity, once physics is the domain, as a single jump; I ask whether the
analogous capacity in mathematics has the same unity, or whether it decomposes once
mathematical practice is examined on its own terms.

My position is that it decomposes, and that treating it as a single capacity - in
either direction, as uncritical optimism about AI mathematical ability or as blanket
skepticism about AI creativity - is a mistake. I identify four mechanistically distinct
modes, not offered as an exhaustive list, by which mathematics has historically produced
new concepts, axioms, and connections: \emph{reflexive} mathematics, which formalizes an existing mathematical or
logical practice by taking the practice itself as an object of study; \emph{analogical}
mathematics, which imports structure discovered in the sciences and repurposes it
inside pure mathematics; \emph{problem-driven} mathematics, in which new concepts arise
as a byproduct of solving it; and the \emph{bridging of distant domains}, in
which value comes specifically from connecting fields with no obvious prior relationship.
Cutting across all four, in a way Section~\ref{sec:background} situates precisely, is a
further distinction developed through the case of conjecture-formation: a claim can be
reached because a pattern was noticed and generalized, or it can instead be adopted as a
target because its truth would deliver a separately wanted result, with its truth
investigated only afterward. For each mode, and for this further axis, I ground the historical
case in an architecture-level account of why current transformer-based systems, which
operate substantially by recombining and searching over a learned library of existing
moves, succeed or fail at it. My central claim is that these mechanisms are likely
\emph{non-substitutable}: strength in the modes current systems already perform well -
existential search, in-space deduction, and straightforward cross-domain transfer - does
not transfer to the modes that require inventing a genuinely new conceptual primitive
rather than a new combination of old ones. Given recombination over a fixed candidate
type, this is what a system cannot reach at all, not merely what would take longer - a
completeness claim, not an efficiency one. The risk sits in that premise, not its
consequence: whether current systems really are recombination engines is defended from an
observed pattern in Section~\ref{sec:problematic} onward, not proved outright.

The remainder of the paper proceeds as follows. \Cref{sec:background} recalls the
Peircean apparatus this paper shares with \citet{zahavy2026jump} and states the
domain-specific question I ask of it. \Cref{sec:taxonomy} develops the four modes and the
further motivational axis in turn. \Cref{sec:unifying} states the non-substitutability
thesis directly. \Cref{sec:migration} argues that mathematical value is migrating toward
the modes current systems cannot yet perform, as the cost of producing and verifying
proofs falls with the improving performance of AI systems at generating them.
\Cref{sec:alternative} takes up credible objections to this position, and
\Cref{sec:conclusion} closes by returning to the sentence in \citet{zahavy2026jump} this
paper set out to answer.

\section{The Peircean frame}
\label{sec:background}

Peirce's classical account of inference divides it according to the structural
permutation of a rule, a case, and a result. Deduction (rule + case $\to$ result) applies
a general rule to a particular case to derive a result with certainty; induction (case +
result $\to$ rule) generalizes a rule from an accumulation of cases and their results; and
abduction (rule + result $\to$ case) infers the case, or a new rule, that would best
explain an otherwise surprising result~\citep{peirce1934collected}. \citet{zahavy2026jump}
adopts this trichotomy to argue that large language models have mastered induction, as
the statistical compression of data, and are rapidly mastering deduction, as the formal
verification of theorems from given premises, but lacks a mechanism for abduction: the
generation of the axioms from which deduction can proceed at all. His case study is
Einstein's invention of general relativity, where the relevant axioms - chiefly the
equivalence principle - were reached not by observing a body of data (little of the
relevant kind yet existed) nor by deducing them from prior theory (they were themselves
the premises), but by what he calls manipulative abduction: an embodied thought
experiment that grounded a new axiom in simulated physical sensation. Although the case
study is drawn from physics, the claim is not confined to it: his own conclusion states
that \textit{``the necessity of the Abductive Jump remains universal,''} varying only in \textit{``the
nature of the simulation,''} which must be \textit{``adapted to the ontology of the
discipline''}-for physics, the world; for mathematics, \textit{``the abstract landscape of formal
systems''}~\citep{zahavy2026jump}.

Zahavy's apparatus is a starting point, not a foundation: none of induction, deduction, or
abduction alone accounts for the origin of new mathematical content, so his
question - where do the new rules come from - is the right one to ask of mathematics
too.
What follows is not a derivation of four sub-species inside his trichotomy, but an
examination of mathematical practice on its own terms, which does not support his claim
that the jump is one capacity, universal in kind, varying only in substrate. New
mathematical rules arise, historically, along several structurally distinct routes,
triggered by different kinds of input - an existing practice to formalize, an external
analogy to import, a problem's own constraints, a distant domain's latent structure.
Whether each route is properly a species of Peircean abduction, or simply shares its
rough shape, is left open; Sections~\ref{sec:reflexive}--\ref{sec:bridging} develop the
four routes regardless, extending Zahavy's question rather than rejecting it.
Section~\ref{sec:desire} then develops a further, orthogonal distinction - not a fifth
route, but a question about what motivates the pursuit of any of the four: whether it is
triggered by an observed pattern or by a separately held goal.

\section{A taxonomy of mathematical meaning-making}
\label{sec:taxonomy}

\subsection{Reflexive mathematics}
\label{sec:reflexive}

Some mathematics formalizes not some external subject matter but an existing
mathematical or logical practice, taking the practice itself as the object of
study - what I call \emph{reflexive} mathematics. Turing's 1936 paper is the paradigm case: it does
not formalize an external phenomenon but the practice of a human \textit{``computer''} mechanically
following a fixed rulebook, and it is this reflexive move, not any subsequent theorem,
that originates the Turing machine~\citep{turing1936computable}. Boole's \emph{Laws of
Thought} is an earlier and more explicit instance: Boole states directly that he is
building a mathematical model of the mental operations of reasoning, not merely a notation
for propositions already agreed upon, reducing connectives such as \textit{``and,'' ``or,''} and
\textit{``not''} to algebraic operations on truth values - the origin of the word \textit{``Boolean''} now
embedded in every programming language~\citep{boole1854laws}. Gentzen's natural deduction
is a third: it supplies a pair of rules for each logical connective, one to introduce it
and one to eliminate it (for instance, one rule concludes \textit{``$A \wedge B$''} from \textit{``$A$''} and
\textit{``$B$''} held separately, and another recovers \textit{``$A$''} from \textit{``$A \wedge B$''}), designed, in
Gentzen's own words, to come as close as possible to the informal moves mathematicians
already make when they argue, and its introduction/elimination structure remains the
backbone of the proof systems used in modern interactive theorem provers such as
Lean~\citep{gentzen1935untersuchungen}. G\"odel's arithmetization of syntax, which turns
the practice of proving into an object inside arithmetic itself, extends the same move to
a second order~\citep{godel1931formal}.

Whether large language models can perform this mode is not settled by asking whether they
can introspect, a question on which the literature is already substantial. Fine-tuned to
predict their own behavior in hypothetical scenarios, language models
outperform other models at predicting them, but only on simple, in-distribution
tasks~\citep{binder2024looking}. Concept-injection experiments, in which a vector is
inserted into a model's residual stream mid-task, find that a frontier model can sometimes
detect and name the injection, but unreliably~\citep{anthropic2025introspective}; a direct
follow-up shows that this particular binary detection paradigm is largely a methodological
artifact of a global bias toward affirmative answers, while a narrower, more robust
capacity survives in relative judgments - localizing which of several sentences was
perturbed, and comparing the strength of two perturbations, though confined to
early-layer injections and collapsing to chance thereafter~\citep{hahami2025detecting}.
Introspection, where it exists at all in these systems, is partial and layer-dependent,
not a unified transparent readout.

This should not be read as a point in favor of humans by contrast. Human introspection is
not a transparent readout either: people confidently report causes for their own behavior
that demonstrably were not the actual causes~\citep{nisbett1977telling}. Neither side has
raw access to its own mechanism, and the disanalogy between them is subtler than it first
appears. In fact, Turing's own move did not require phenomenal, first-person transparency:
the object he formalized, a human clerk following a rulebook, is third-person and
behaviorally observable, not introspectively private. Language models, trained on an
immense record of human mathematical practice, arguably have more systematic third-person
exposure to that practice than any single human mathematician does. Data access, then, is
probably not the real bottleneck to reflexive mathematics.

The more likely bottleneck is \emph{object selection}: nothing in ordinary training or
inference makes an existing symbolic or cognitive practice salient as a thing worth
turning into a formal object in its own right, absent someone external pointing at it.
This is the structural feature shared by all four case studies above, not a peculiarity
of Turing's alone: in each, what got selected for formalization was the practice
itself - computing, for Turing; reasoning, for Boole; the structure of argument, for
Gentzen; proof-construction, for G\"odel - rather than some external subject matter the
practice happened to be used on. This reframes reflexive mathematics as an instance of a
more general concept-creation problem - which practice gets selected as worth
formalizing - rather than a separately blocked introspective faculty.
One candidate account of how such selection happens is retrospective: a candidate
formalization is proposed, and its fruitfulness, how much rich, unifying structure follows
from it, validates the choice after the fact. If this is right, object selection resembles
a search-and-score loop, propose a formalization and evaluate it by downstream
theorem-yield, much closer to what evolutionary and search-based AI systems already do
than to a mysterious human faculty, and it directly extends \citeauthor{zahavy2026jump}'s
own suggestion that mathematics' sense experience may be grounded in \textit{``generality or
minimality.''} This is a genuine partial rescue, but I do not think it closes the gap,
because it addresses only half the problem.
Fruitfulness-scoring evaluates a candidate once one exists; it says nothing about how the
candidate's \emph{type} arises in the first place. Systems that propose and score candidate
mathematical objects, such as AlphaEvolve and FunSearch, search a space whose type is
already fixed by the problem statement - find a set, a sequence, a construction of an
already-named kind~\citep{novikov2025alphaevolve,romeraparedes2024mathematical}. Turing's
case makes this clearest: he did not search a pre-given space of \textit{``possible
formalizations of computation''} and score its members; he first had to treat \textit{``the
practice of a human calculating''} as a legitimate kind of object to formalize, before any
scoring loop could begin, and the same holds for Boole's choice of reasoning, Gentzen's
choice of argument, and G\"odel's choice of proof as objects worth formalizing in their
own right. The scoring loop is the easy part; picking what to run it over is what
reflexive mathematics actually demands, and that is where current systems stall.

\subsection{Analogical mathematics}
\label{sec:analogical}

A second mode creates new concepts by importing structure discovered in the sciences
into pure mathematics, where it is repurposed to answer questions with no dependence on
the domain it came from; I call it \emph{analogical} mathematics. The physical sciences supply the
clearest running examples below, but nothing in the mechanism restricts the source to
physics specifically: biology, for instance, is just as plausible a source domain in
principle. A familiar instance of the move is Fourier's decomposition of heat conduction
into sinusoidal modes, introduced to solve a specific physical problem and subsequently
generalized into harmonic analysis, a foundational area of pure mathematics with no
remaining dependence on heat or on any other physical process. The clearest
\emph{sustained} example, with the longest lineage of major results built on the imported
concept, is entropy. Ergodic theory had, by the 1950s, an unsolved classification
problem: given two Bernoulli shifts, decide whether they are measure-theoretically
isomorphic, with no known invariant able to tell most pairs apart. Kolmogorov and Sinai's
move was to notice that the statistical entropy of Boltzmann and Gibbs, redefined
appropriately, behaves as exactly such an invariant for abstract measure-preserving
dynamical systems~\citep{kolmogorov1958new,kolmogorov1959entropy,sinai1959notion}. Ornstein
closed the loop in 1970, proving the invariant complete: two Bernoulli shifts are
isomorphic exactly when their Kolmogorov--Sinai entropies
agree~\citep{ornstein1970bernoulli}. Decades later the same invariant was still doing new
work: Bowen extended it to actions of countable sofic
groups, resolving isomorphism questions the classical invariant could not
reach~\citep{bowen2010measure}, showing that the imported concept is still generative
decades after the original analogy was drawn.

Of the four modes in this taxonomy, this is the one I am most conditionally optimistic
about. \citet{zahavy2026jump} argues that physically grounded, actionable world
models - systems that support intervention on a simulation, such as Genie~\citep{bruce2024genie},
rather than merely predicting its next frame - could supply the sensory grounding
current language models lack. We can extend this specifically to analogical mathematics:
a world model that has
extracted genuine physical understanding, not merely pixel-level prediction, could
plausibly extract mathematical objects like entropy from that understanding and repurpose
them for questions with no physical content. Concretely: repeated counterfactual
intervention on a physical system - turning a dial, running the same process forward
from different initial conditions - exposes some quantity as invariant, or as changing
only in one direction, independently of the specific physical instantiation being
manipulated. Some
such pattern of invariance becoming \emph{salient} as a candidate for abstraction is
plausibly what let Kolmogorov and Sinai treat a physical quantity as the right analogy for
an abstract isomorphism invariant, whatever the specific route by which they themselves
arrived at it; reproducing that kind of salience, not grounding in general, is what a
world model would need to support for analogical mathematics to become reachable by an AI
system.

\subsection{Problem-driven mathematics}
\label{sec:problematic}

A third source of new concepts is neither a formalized practice nor an imported
structure, but a specific problem solved, with the concepts arising only as a byproduct:
\emph{problem-driven} mathematics. This mode splits along a line that matters for what current systems
can do with it. In its \emph{existential} or constructive form, the goal is a witness: a
counterexample, or a construction meeting some extremal constraint. A great deal of what
current AI-assisted mathematics has actually produced is of this shape: FunSearch and
AlphaEvolve's headline results - improved constructions for the cap set problem, better
bounds on kissing numbers, faster matrix-multiplication
algorithms - are almost uniformly existential~\citep{romeraparedes2024mathematical,novikov2025alphaevolve}.
In its \emph{universal} or structural form, by contrast, the goal is a new organizing
theory that reorganizes an entire domain rather than answering one existential question,
and no accumulation of witnesses could have produced it. Abel had shown by 1824 that no
general formula in radicals solves the quintic, but this left open why some specific
quintics are solvable and others are not - a question no list of examples could settle.
Galois's answer was to associate to each polynomial a group of permutations of its roots
and to show that solvability by radicals corresponds to a specific structural property of
that group. The result was not a new technique for solving equations but a new kind of
object, group theory itself, under which the whole question could be redescribed and
closed. Non-Euclidean geometry answers a structurally similar resistance - centuries of
failed attempts to prove the parallel postulate from the other four - with a comparable
move: a new kind of object, a consistent geometry in which the postulate simply fails,
rather than a witness. Both cases are shaped like the abductive jump of
\citet{zahavy2026jump}: a persistent, surprising resistance resolved not by deduction or
generalization, but by inferring an entirely new explanatory apparatus.

The architectural story developed throughout this taxonomy is a conjecture inferred from
an observed pattern, not an established mechanistic fact. Essentially every headline
result from autonomous, self-directed AI mathematical discovery - where the system
chooses what to search for, not merely completes a proof already posed - is existential;
separately, a mechanistic study of language models solving arithmetic finds them relying
on a sparse set of narrow, interpretable heuristics rather than a general
procedure~\citep{nikankin2025arithmetic}, indirect evidence, at a much smaller scale, for
the same picture. Taking the pattern at face value: a transformer's attention mechanism,
refined by search or reinforcement learning over candidate proof steps, plausibly
recombines a learned repertoire of existing moves. This repertoire need not be formalized: alongside named lemmas, tactics, and
constructions already seen in training or discovered by search, it includes looser
fragments of informal reasoning that a human reader would not necessarily isolate as a
standalone step within the surrounding proof. Recombination, whether over formal or
informal moves, is naturally shaped like witness-finding: search a combinatorial space of
known gadgets for one instance satisfying a constraint, exactly what an existential claim,
or a counterexample to a universal one, asks for. It is not naturally shaped like
structural reorganization, which asks not for an instance of a known kind of object but
for a new kind of object under which the whole domain can be redescribed.

What makes a candidate object genuinely new, rather than a fresh assembly of existing
pieces, is not something this paper resolves: every new definition is, at some level,
built from existing primitives. One candidate criterion, offered as a position rather
than a proof: a new counterexample is typically long and specific to the instance that
produced it; a new concept is typically short, and opens results not provable before it
existed. The criterion is retrospective, not a test a system's output could pass on the
spot - newness becomes clear only once downstream consequences do - but it still
predicts something: any AI-discovered concept later confirmed genuine should turn out
short and reused, not a longer witness mistaken for one. \citet{gowers2026llms}
independently draws the same line from live experience with frontier systems: search
breadth beats any mathematician, pruning judgment does not, and genuine progress is
measured by an equally retrospective standard - a technique unremarkable in hindsight
but \textit{``difficult to stumble on by accident.''} His observations stay entirely
inside problem-driven mathematics, though, and say nothing about whether the same line
holds for reflexive formalization or bridging distant domains - the generality claim
this paper makes and his does not. That \citet{zahavy2026jump} and \citet{gowers2026llms}
converge on the same split from opposite directions - philosophy of mind and live
experimentation - is evidence it tracks something real, not an artifact of either
paper's framing.

\subsection{Bridging distant domains}
\label{sec:bridging}

A fourth mode's value comes specifically from connecting two bodies of
mathematics - entire fields, but just as often two specific objects, theorems, or
results within or across them - with no obvious prior relationship: the
\emph{bridging} of distant domains, historically one of the most prized forms of
mathematical unification. Part of why it is
prized is a matter of division of labor: mastering one field well enough to contribute to
it already absorbs most of a working mathematician's career, so mastering several fields
well enough to notice that a bridge between them is even possible is correspondingly
rarer, and unifying results have carried outsized prestige in part because so few people
are ever positioned to attempt them. This constraint does not obviously bind an AI system
trained on the literature of many fields at once, unbottlenecked by a single lifetime of
specialization - one reason, developed below, that surfacing a candidate connection is
plausibly closer to reach than constructing one where none yet exists. This mode also
splits usefully into two cases. In the first, a translation dictionary between the two
domains already exists, or becomes available once the right quantities are computed on
both sides; connecting them is then a matter of recognizing an already-available plug. In
the second, no such interface exists, and bridging the domains requires inventing new
intermediate machinery from scratch. \citet{davies2021advancing} give a clean, concrete
instance of the difference between finding a plug and filling a gap. A supervised model
trained to relate algebraic invariants of knots to their hyperbolic-geometry invariants
(and, separately, a structure connecting representation theory and combinatorics) surfaced
a strong statistical correlation between quantities already computable on both sides of
each divide - exactly the \textit{``plug''} case. But it took a human mathematician to construct
the theorem explaining why the correlation held, the actual bridging move in the sense
that matters for the harder case. Current systems look plausibly capable of the first case
and structurally unequipped for the second, for the same architectural reason given
throughout this section: recombining what is already computable is a search problem;
building the interface where none exists is a concept-creation problem.

A related mechanism worth distinguishing from bridging, though space precludes
developing it as its own case here, is \emph{unification}: rather than an interface
between two existing structures that remain standing on either side of it, unification
defines a single, more general object into which several existing objects particularize
as special cases, sometimes from domains with no prior relationship at all - group
theory's absorption of the many symmetry groups already in separate use, or category
theory's later absorption of structures across most of mathematics, are paradigm
instances. Where bridging adds a connection, unification typically dissolves the
connected structures into instances of something new, which can have the further effect,
when it reaches across domains, of transferring techniques from one instance to all the
others. Whether this is truly independent of the constructive move behind
Sections~\ref{sec:problematic}--\ref{sec:desire}, or another guise of it, is left to
future work.

\subsection{Observation-driven versus goal-driven conjecture}
\label{sec:desire}

As anticipated in Section~\ref{sec:background}, the axis this section develops is
sharpest in the case of conjecture-formation: a conjecture is \emph{observation-driven}
when it is reached inductively, by noticing a pattern across many cases and
generalizing it, and
\emph{goal-driven} when a claim is instead adopted as a target because, if true, it would
deliver a separately wanted result, with its truth investigated only afterward. The
Collatz conjecture is observation-driven in exactly this sense: it generalizes from an
accumulation of computed trajectories, with no external motivating goal beyond the pattern
itself. The route to Fermat's Last Theorem through the Frey curve is goal-driven in the
contrasting sense. Frey proposed that a specific elliptic curve, built from a hypothetical
counterexample to Fermat's Last Theorem, would fail to be modular~\citep{frey1986links};
Ribet proved the corresponding epsilon conjecture, confirming that this failure would
indeed follow~\citep{ribet1990modular}; and Wiles, together with Taylor, proved enough of
the modularity conjecture to complete the theorem~\citep{wiles1995modular,taylorwiles1995ring}.
The entire program of proving modularity for this specific curve was pursued because it
would deliver Fermat's Last Theorem, not because of independent pattern-noticing about
elliptic curves - a goal-driven conjecture in a sense Collatz never was.

This axis cuts across the same depth distinction developed in
Sections~\ref{sec:reflexive}--\ref{sec:bridging}, rather than resolving to a single
verdict. In its shallow, in-space form, goal-driven reasoning is goal-directed subgoal
search within an already specified space of lemmas and constructions - essentially
backward chaining: work back from the desired conclusion, propose an intermediate lemma,
and check whether existing tactics close the resulting subgoal. This is the same
recombination mechanism described throughout this taxonomy, applied under an explicit
goal rather than free exploration, which is why it is close to what
reinforcement-learning-guided theorem provers, already searching over exactly this kind
of fixed space via expert iteration, do reasonably well. In its deep, out-of-space form,
exemplified by Frey's move, the target claim requires constructing a genuinely new
mathematical object as the connecting device - a specific elliptic curve built from a
hypothetical counterexample, whose relevance had no precedent to search over - not
selecting among known lemmas, before its truth is even known. Nothing in the space of
previously considered elliptic curves would have flagged this particular construction as
worth proposing; the construction itself, and not merely its eventual truth, was the
creative act. The first case looks reachable by current search-augmented systems because
it searches an already-fixed candidate type, exactly the sense in which
Section~\ref{sec:reflexive} found the evaluation half of reflexive mathematics
tractable; the second is stuck for the same reason every hard case in this taxonomy is
stuck - there was no space of candidates to search until Frey built one.

\section{Non-substitutability of mechanisms}
\label{sec:unifying}

The four modes developed in Section~\ref{sec:taxonomy}, together with the further
motivational axis developed alongside them, are not independent paths to the same
destination. My position is that they are distinct routes into what a mathematician would
recognize as concept space, and that they are probably not mutually substitutable:
strength in one does not transfer to the others. The claim is about reachability, in
other words, rather than speed: certain mathematics sits outside what the architecture
current systems share can produce at all, regardless of how much longer a slower search
might be given.

The same underlying mechanism explains both the pattern of successes and the pattern of
gaps across every section above. Attention-based recombination over a learned library of
existing moves, refined by search or reinforcement learning, is well suited to tasks whose
candidate space has a fixed, already-specified type: deduction from given axioms, the
existential half of problem-driven mathematics (Section~\ref{sec:problematic}), the plug
half of bridging distant domains (Section~\ref{sec:bridging}), and the shallow half of
goal-driven conjecture (Section~\ref{sec:desire}). It is not well suited to tasks that
require inventing the candidate type itself, rather than searching within one already
given: reflexive mathematics (Section~\ref{sec:reflexive}), the structural half of
problem-driven mathematics, the gap-filling half of bridging distant domains, and the deep
half of goal-driven conjecture. Analogical mathematics (Section~\ref{sec:analogical})
sits apart from this pattern precisely because it is conditional on a different kind of
architectural progress, grounded world models, rather than on scaling the recombination
mechanism that governs the rest of the taxonomy. Concept creation, in the sense of
establishing a new stable primitive rather than a new combination of existing
ones - close to \citet{boden2004creative}'s transformational creativity, against the
combinational and exploratory creativity recombination already covers - is therefore
not one item on a longer list of missing capabilities; it is the single recurring
bottleneck that the taxonomy in Section~\ref{sec:taxonomy} decomposes into four
historically distinguishable mechanisms, cut across in every case by the further question
of what motivated their pursuit. Borrowing the vocabulary from
Section~\ref{sec:background} without depending on it: every intractable half has the
rough shape of abduction, inferring something new rather than applying or generalizing
what is given; every tractable half has the rough shape of deduction or induction over an
already-fixed space. The claim survives without the vocabulary: one missing capacity in a
single physics case study recurs, distributed across mathematical practice, as several
distinct obstacles with a family resemblance.

Whether the structural half of problem-driven mathematics, the gap-filling half of
bridging, and the deep half of goal-driven conjecture are genuinely separate mechanisms, or one
constructive move under three different triggers, is a question the historical cases
alone probably cannot settle, and this paper does not try to. That caps how strong the
taxonomic claim can be; what survives the cap is narrower, but real: a resistant problem,
an interface-less pair of domains, and a goal with no existing lemma are recognizably
different situations to notice, whatever their deeper relationship turns out to be.
Organizing around what must be perceived to trigger a construction, rather than around
its deepest unity, is what makes the evaluation proposal in Section~\ref{sec:migration}
useful either way - and why bridging keeps its own
section regardless of how that further question comes out.

\section{Where mathematical value is migrating}
\label{sec:migration}

Section~\ref{sec:intro} opened with Tao's diagnosis that the field is moving from a
shortage of proofs to an overwhelming abundance of them~\citep{tao2026mathematics}. If
Section~\ref{sec:unifying} is right that this abundance is concentrated in modes built
from recombination and search - deduction, existential construction, straightforward
transfer - rather than in modes that require new conceptual primitives, then the
practical upshot is not merely that the cost of proof is falling in general. It is that
mathematical value, both the comparative advantage of individual mathematicians and the
allocation of the field's collective effort, should migrate toward precisely the modes
Section~\ref{sec:taxonomy} identifies as currently out of reach: reflexive formalization
of practice, analogical import pending grounded world models, structural
reorganization, gap-filling between domains, and the deep, goal-driven form of conjecture.

This migration is not unambiguously good news; it runs in two directions at once. On one
branch, less costly proof frees mathematicians
from the labor of verification to spend more of their attention on exactly the
concept-level work this taxonomy identifies as still scarce. On the other, Tao's own
warning is that \textit{``the inherently ungrounded nature of generative AI, combined with the
financial incentives of AI companies, makes the use of these tools particularly vulnerable
to''} Goodhart's law~\citep{tao2026mathematics}: once a measure such as proof count or
benchmark performance becomes a target, it stops being a good measure. A field that
already produces proofs of uninteresting results because proof itself is automatically
valued, independent of whether the result is interesting, is a dynamic visible before any
of the recent AI systems existed; driving the cost of proof down further could just as
easily flood the literature with more of the same rather than freeing anyone's attention
for anything else. Both branches follow from the same premise - that recombination-shaped
work is becoming abundant - and which one dominates in practice depends on incentives
this paper does not attempt to redesign, only to name.

This is a recommendation, not only a diagnosis. If AI mathematical systems' value is
concentrated in a specific subset of Section~\ref{sec:taxonomy}'s modes, a single
aggregate score for \textit{``mathematical ability''} measures the wrong thing: it cannot
distinguish strength at existential search from a genuinely new mode of concept creation,
and under Goodhart's law an aggregate invites optimizing the number rather than the
capacity itself. Benchmark designers and evaluators should report performance broken out
by mode, rather than folding reflexive, analogical, problem-driven, and bridging tasks, at
any point on the observation/goal axis, into one figure - making the taxonomy
falsifiable by disaggregation, not just defensible as philosophy.

\section{Alternative views}
\label{sec:alternative}

A credible alternative holds that recombination is not fundamentally limited, only
under-scaled: enough compute, better search scaffolding, or expert-iteration training like
AlphaProof's might reach the modes called structurally out of reach here. But scaling
recombination searches a candidate type more thoroughly; it does not expand what counts as
one, the claim in Section~\ref{sec:unifying} - a difference in kind, not degree. This
needs a case where scaled search alone produced a genuinely new primitive rather than a
better search of an old one. None exists that I know of, but I hold the claim as a bet
that could lose, not a matter of principle. A related version is that prompting might
already elicit the missing capacity, a concern raised publicly against the closest prior
work in this vein\footnote{Reviews of the closest prior work in this vein post publicly on
OpenReview post-decision; this concern appears in a review of \citet{zahavy2026jump}.}:
prompting works within an existing
symbolic space, combining what is already there, which is why it does not reach outside
it.

The next objection runs against my optimism rather than my pessimism:
Section~\ref{sec:migration} argued cheaper proof could free mathematicians for
concept-level work, but conceded, following Tao, that it could just as easily degrade the
field's signal-to-noise with more correct, uninteresting results. I regard this as a
live, unresolved tension, not something the taxonomy resolves.

Section~\ref{sec:reflexive} faces another objection: if fruitfulness-scoring rescues the
evaluation half of object selection, perhaps it rescues generation too, undermining the
claim that reflexive mathematics stays out of reach. Doubtful, because every
search-and-score system known searches a space whose type the problem statement already
fixed; no fruitfulness criterion tells a system which space to search in the first place.
This is an empirical bet, not a first principle, and it loses the moment a system proposes
a new category of object, not just a new member of an old one, later vindicated by its
consequences.

The diagnosis in Section~\ref{sec:reflexive} that data access
is not the bottleneck to reflexive mathematics is questioned too: language models are trained on an
enormous third-person record of human mathematical practice, so perhaps that exposure
already suffices, and object selection is an unnecessary further bottleneck. My response
is that exposure to a practice and treating it as a salient object of formalization are
different achievements: a system trained on transcripts of computation does not thereby
select computation itself, rather than any other regularity in its training data, as
worth axiomatizing. The disagreement is about salience, not sufficiency, and I think the
distinction survives.

One last objection concerns significance, not truth: that this taxonomy merely illustrates
Zahavy's abductive jump in a new domain. It corrects his claim rather than instantiating
it. His conclusion asserts that \textit{``the necessity of the Abductive Jump remains
universal''}; the position here is that this holds only at the level of Peirce's most
general category, and is radically underspecified at the level that matters for building
or evaluating a system, since what triggers the inference differs by mechanism - a
practice made salient, a structure made available through grounded simulation, or a
resistant problem, missing interface, or unreachable goal made salient as such. Moving
from a name for the missing capacity to a falsifiable account of what must be supplied,
case by case, is what a single case study could not have produced.

\section{Conclusion}
\label{sec:conclusion}

This paper asked how mathematical creativity should be assessed once large language
models are capable of producing correct, verified mathematics at scale, and argued that
the question is currently underspecified. Mathematical concept-creation decomposes into
four mechanistically distinct modes - reflexive, analogical, problem-driven, and the
bridging of distant domains - cut across by a further axis distinguishing conjecture
reached from observation and conjecture pursued out of a separately held goal, and I have
argued that these mechanisms are probably non-substitutable, so that a system's
competence in the modes shaped by recombination and search does not license any inference
about its competence in the modes that require inventing a new conceptual primitive. This
is why I think blanket claims about AI and mathematical ability, in either direction, are
mistaken: the right question is never simply whether a system can do mathematics, but
which of these mechanisms a given claim of capability or incapacity is actually about.

This paper began from a single sentence left open by \citet{zahavy2026jump}, who
correctly declined to force his physics-derived account of the abductive jump onto a
domain he had not examined, suggesting only that mathematics' sense experience \textit{``may be
grounded in high-dimensional topology or have other goals such as generality or
minimality.''} I have tried to take that sentence seriously rather than merely cite it: not
by proposing a second jump to sit beside his, but by asking whether the very idea of a
single jump survives contact with how mathematics has actually produced new axioms,
concepts, and conjectures. My answer is that it does not, and that the taxonomy developed
here is one way, not the only one, of taking his open question further.

Two threads are left underdeveloped for space, not substance:
operationalizing object selection as the bottleneck to reflexive mathematics
(Section~\ref{sec:reflexive}); and proof's rise as mathematics' currency, sketched briefly
through Hardy and Ramanujan. Neither changes the position argued; the taxonomy
itself is incomplete, one contribution among others.

\bibliography{references}

@book{peirce1934collected,
  author    = {Peirce, Charles Sanders},
  title     = {Collected Papers of Charles Sanders Peirce},
  publisher = {Harvard University Press},
  year      = {1934}
}

@book{boden2004creative,
  author    = {Boden, Margaret A.},
  title     = {The Creative Mind: Myths and Mechanisms},
  edition   = {2nd},
  publisher = {Routledge},
  year      = {2004}
}

@inproceedings{zahavy2026jump,
  author    = {Zahavy, Tom},
  title     = {Position: {LLMs} Can't Jump},
  booktitle = {Proceedings of the 43rd International Conference on Machine Learning (ICML)},
  year      = {2026}
}

@misc{bruce2024genie,
  author       = {Bruce, Jake and Dennis, Michael and Edwards, Ashley and Parker-Holder, Jack and Shi, Yuge and Hughes, Edward and Lai, Matthew and Mavalankar, Aditi and Steigerwald, Richie and Apps, Chris and Aytar, Yusuf and Bechtle, Sarah and Behbahani, Feryal and Chan, Stephanie and Heess, Nicolas and Gonzalez, Lucy and Osindero, Simon and Ozair, Sherjil and Reed, Scott and Zhang, Jingwei and Zolna, Konrad and Clune, Jeff and de Freitas, Nando and Singh, Satinder and Rockt{\"a}schel, Tim},
  title        = {Genie: Generative Interactive Environments},
  year         = {2024},
  howpublished = {arXiv:2402.15391}
}

@article{turing1936computable,
  author  = {Turing, Alan M.},
  title   = {On Computable Numbers, with an Application to the {Entscheidungsproblem}},
  journal = {Proceedings of the London Mathematical Society},
  series  = {2},
  volume  = {42},
  pages   = {230--265},
  year    = {1936}
}

@book{boole1854laws,
  author    = {Boole, George},
  title     = {An Investigation of the Laws of Thought, on Which Are Founded the Mathematical Theories of Logic and Probabilities},
  publisher = {Macmillan and Co.},
  address   = {London},
  year      = {1854}
}

@article{gentzen1935untersuchungen,
  author  = {Gentzen, Gerhard},
  title   = {Untersuchungen {\"u}ber das logische {Schlie{\ss}en}. {I}, {II}},
  journal = {Mathematische Zeitschrift},
  volume  = {39},
  pages   = {176--210, 405--431},
  year    = {1935}
}

@article{godel1931formal,
  author  = {G{\"o}del, Kurt},
  title   = {{\"U}ber formal unentscheidbare {S}{\"a}tze der {Principia Mathematica} und verwandter {Systeme I}},
  journal = {Monatshefte f{\"u}r Mathematik und Physik},
  volume  = {38},
  pages   = {173--198},
  year    = {1931}
}

@article{binder2024looking,
  author  = {Binder, Felix J. and Chua, James and Korbak, Tomek and Balesni, Marius and Rager, Cameron and Kori, Chen and Meek, Nathan and Evans, Owain},
  title   = {Looking Inward: Language Models Can Learn About Themselves by Introspection},
  journal = {arXiv preprint arXiv:2410.13787},
  year    = {2024}
}

@misc{anthropic2025introspective,
  author       = {{Anthropic}},
  title        = {Emergent Introspective Awareness in Large Language Models},
  year         = {2025},
  howpublished = {\url{https://transformer-circuits.pub/2025/introspection/index.html}}
}

@article{hahami2025detecting,
  author  = {Hahami, Ely and Sinha, Ishaan and Jain, Lavik and Kaplan, Josh and Hahami, Jon},
  title   = {Detecting the Disturbance: A Nuanced View of Introspective Abilities in {LLMs}},
  journal = {arXiv preprint arXiv:2512.12411},
  year    = {2025}
}

@inproceedings{nikankin2025arithmetic,
  author    = {Nikankin, Yaniv and Reusch, Anja and Mueller, Aaron and Belinkov, Yonatan},
  title     = {Arithmetic Without Algorithms: Language Models Solve Math With a Bag of Heuristics},
  booktitle = {International Conference on Learning Representations (ICLR)},
  year      = {2025},
  note      = {arXiv:2410.21272}
}

@article{nisbett1977telling,
  author  = {Nisbett, Richard E. and Wilson, Timothy D.},
  title   = {Telling More Than We Can Know: Verbal Reports on Mental Processes},
  journal = {Psychological Review},
  volume  = {84},
  number  = {3},
  pages   = {231--259},
  year    = {1977}
}

@article{kolmogorov1958new,
  author  = {Kolmogorov, Andrei N.},
  title   = {New Metric Invariant of Transitive Dynamical Systems and Endomorphisms of {Lebesgue} Spaces},
  journal = {Doklady Akademii Nauk SSSR},
  volume  = {119},
  number  = {5},
  pages   = {861--864},
  year    = {1958}
}

@article{kolmogorov1959entropy,
  author  = {Kolmogorov, Andrei N.},
  title   = {Entropy per Unit Time as a Metric Invariant of Automorphisms},
  journal = {Doklady Akademii Nauk SSSR},
  volume  = {124},
  pages   = {754--755},
  year    = {1959}
}

@article{sinai1959notion,
  author  = {Sinai, Yakov G.},
  title   = {On the Notion of Entropy of a Dynamical System},
  journal = {Doklady Akademii Nauk SSSR},
  volume  = {124},
  pages   = {768--771},
  year    = {1959}
}

@article{ornstein1970bernoulli,
  author  = {Ornstein, Donald S.},
  title   = {Bernoulli Shifts with the Same Entropy Are Isomorphic},
  journal = {Advances in Mathematics},
  volume  = {4},
  number  = {3},
  pages   = {337--352},
  year    = {1970}
}

@article{bowen2010measure,
  author  = {Bowen, Lewis},
  title   = {Measure Conjugacy Invariants for Actions of Countable Sofic Groups},
  journal = {Journal of the American Mathematical Society},
  volume  = {23},
  number  = {1},
  pages   = {217--245},
  year    = {2010}
}

@article{davies2021advancing,
  author  = {Davies, Alex and Veli{\v{c}}kovi{\'c}, Petar and Buesing, Lars and Blackwell, Sam and Zheng, Daniel and Toma{\v{s}}ev, Nenad and Tanburn, Richard and Battaglia, Peter and Blundell, Charles and Juh{\'a}sz, Andr{\'a}s and Lackenby, Marc and Williamson, Geordie and Hassabis, Demis and Kohli, Pushmeet},
  title   = {Advancing Mathematics by Guiding Human Intuition with {AI}},
  journal = {Nature},
  volume  = {600},
  pages   = {70--74},
  year    = {2021}
}

@article{frey1986links,
  author  = {Frey, Gerhard},
  title   = {Links Between Stable Elliptic Curves and Certain {Diophantine} Equations},
  journal = {Annales Universitatis Saraviensis. Series Mathematicae},
  volume  = {1},
  pages   = {1--40},
  year    = {1986}
}

@article{ribet1990modular,
  author  = {Ribet, Kenneth A.},
  title   = {On Modular Representations of {Gal}$(\overline{\mathbb{Q}}/\mathbb{Q})$ Arising from Modular Forms},
  journal = {Inventiones Mathematicae},
  volume  = {100},
  number  = {2},
  pages   = {431--476},
  year    = {1990}
}

@article{wiles1995modular,
  author  = {Wiles, Andrew},
  title   = {Modular Elliptic Curves and {Fermat}'s Last Theorem},
  journal = {Annals of Mathematics},
  volume  = {141},
  number  = {3},
  pages   = {443--551},
  year    = {1995}
}

@article{taylorwiles1995ring,
  author  = {Taylor, Richard and Wiles, Andrew},
  title   = {Ring-Theoretic Properties of Certain {Hecke} Algebras},
  journal = {Annals of Mathematics},
  volume  = {141},
  number  = {3},
  pages   = {553--572},
  year    = {1995}
}

@article{romeraparedes2024mathematical,
  author  = {Romera-Paredes, Bernardino and Barekatain, Mohammadamin and Novikov, Alexander and Balog, Matej and Kumar, M. Pawan and Dupont, Emilien and Ruiz, Francisco J. R. and Ellenberg, Jordan S. and Wang, Pengming and Fawzi, Omar and Kohli, Pushmeet and Fawzi, Alhussein},
  title   = {Mathematical Discoveries from Program Search with Large Language Models},
  journal = {Nature},
  volume  = {625},
  number  = {7995},
  pages   = {468--475},
  year    = {2024}
}

@misc{tao2026mathematics,
  author       = {Tao, Terence},
  title        = {Mathematics in the Age of {AI}},
  year         = {2026},
  howpublished = {ICM 2026 plenary lecture, \url{https://teorth.github.io/tao-web/slides/age-of-ai-icm-2026.pdf}}
}

@book{hardy1940apology,
  author    = {Hardy, G. H.},
  title     = {A Mathematician's Apology},
  publisher = {Cambridge University Press},
  year      = {1940}
}

@misc{novikov2025alphaevolve,
  author       = {Novikov, Alexander and V{\~u}, Ngoc and Eisenberger, Marvin and Dupont, Emilien and Huang, Po-Sen and Wagner, Adam Zsolt and Shirobokov, Sergey and Kozlovskii, Borislav and Ruiz, Francisco J. R. and Mehrabian, Abbas and Kumar, M. Pawan and See, Abigail and Chaudhuri, Swarat and Holland, George and Davies, Alex and Nowozin, Sebastian and Kohli, Pushmeet and Balog, Matej},
  title        = {{AlphaEvolve}: A Coding Agent for Scientific and Algorithmic Discovery},
  year         = {2025},
  howpublished = {arXiv:2506.13131}
}

@misc{gowers2026llms,
  author       = {Gowers, Timothy},
  title        = {What Sort of Maths Are {LLMs} Good At?},
  year         = {2026},
  howpublished = {Blog post, \url{https://gowers.wordpress.com/2026/08/12/what-sort-of-maths-are-llms-good-at/}}
}
\bibliographystyle{icml2026}

\end{document}